\pdfoutput=1

\documentclass[11pt]{article}

\usepackage[final]{EMNLP2022}

\usepackage{times}
\usepackage{latexsym}

\usepackage{amsmath}
\usepackage{multirow}
\usepackage{booktabs}
\usepackage{graphicx}
\usepackage{subcaption}
\usepackage{tabularx}
\usepackage{bbm}

\newcommand{\hlred}{\colorbox{red!20}}
\newcommand{\hlgreen}{\colorbox{green!20}}
\newcommand{\hlyellow}{\colorbox{yellow!20}}

\usepackage[T1]{fontenc}

\usepackage[utf8]{inputenc}

\usepackage{microtype}

\usepackage{inconsolata}

\usepackage{balance}

\title{AutoCAD: Automatically Generate Counterfactuals for Mitigating Shortcut Learning}

\author{
  Jiaxin Wen$^{1,2*}$, 
  Yeshuang Zhu$^{3}$,
  Jinchao Zhang$^{3}$,
  Jie Zhou$^{3}$,
  Minlie Huang$^{1,2,\dagger}$ \\
  $^1$The CoAI group, Tsinghua University, Beijing, China \\
  $^2$Department of Computer Science and Technology, Tsinghua University, Beijing, China\\
  $^3$Pattern Recognition Center, WeChat AI, Tencent Inc, China \\
  \texttt{wenjx22@mails.tsinghua.edu.cn, aihuang@tsinghua.edu.cn} \\ \texttt{\{yshzhu,dayerzhang,withtomzhou\}@tencent.com}
  }

\begin{document}
\maketitle
\begin{abstract}

Recent studies have shown the impressive efficacy of counterfactually augmented data (CAD) for reducing NLU models' reliance on spurious features and improving their generalizability. However, current methods still heavily rely on human efforts or task-specific designs to generate counterfactuals, thereby impeding CAD's applicability to a broad range of NLU tasks. In this paper, we present AutoCAD, a fully automatic and task-agnostic CAD generation framework. AutoCAD first leverages a classifier to unsupervisedly identify rationales as spans to be intervened, which disentangles spurious and causal features. Then, AutoCAD performs controllable generation enhanced by unlikelihood training to produce diverse counterfactuals. Extensive evaluations on multiple out-of-domain and challenge benchmarks demonstrate that AutoCAD consistently and significantly boosts the out-of-distribution performance of powerful pre-trained models across different NLU tasks, which is comparable or even better than previous state-of-the-art human-in-the-loop or task-specific CAD methods. The code is publicly available at \url{https://github.com/thu-coai/AutoCAD}. 

\end{abstract}

\section{Introduction}

{\let\thefootnote\relax\footnotetext{
$^\dagger$ Corresponding author}
\let\thefootnote\relax\footnotetext{
$^*$ This work was done when Jiaxin Wen was an intern at WeChat AI. }
}

\begin{table}[t]
    \centering
    \resizebox{0.48\textwidth}{!}
    {
        \begin{tabular}{ll}
        \toprule
        \midrule
        \multicolumn{2}{c}{\it{\textbf{Original}}}\\
        \midrule
        \textbf{Premise} & A group of men \hlred{riding bicycles} in a \hlred{line.} \\
        \textbf{Hypothesis} & The men \hlred{riding together.} \\
        \textbf{Label} & \hlred{Entailment} \\
        \midrule
        \midrule
        \multicolumn{2}{c}{\it{\textbf{AutoCAD}}}\\
        \midrule
        \textbf{Premise} & A group of men riding bicycles in a line. \\
        \textbf{Hypothesis} & The men \hlyellow{are professionals.} \\
        \textbf{Label} & \hlyellow{Neutral} \\
        \midrule
        \textbf{Premise} & A group of men riding bicycles in a line. \\
        \textbf{Hypothesis} & The men \hlyellow{ride to work.} \\
        \textbf{Label} & \hlyellow{Neutral} \\
        \midrule
        \textbf{Premise} & A group of men riding bicycles in a line. \\
        \textbf{Hypothesis} & The men \hlgreen{riding horses.} \\
        \textbf{Label} & \hlgreen{Contradiction} \\
        \midrule
        \textbf{Premise} & A group of men \hlgreen{sitting} in a \hlgreen{crowded cafe.} \\
        \textbf{Hypothesis} & The men riding together. \\
        \textbf{Label} & \hlgreen{Contradiction} \\
        \midrule
        \textbf{Premise} & A group of men \hlgreen{riding separately} in a \hlgreen{crowded bus.} \\
        \textbf{Hypothesis} & The men riding together. \\
        \textbf{Label} & \hlgreen{Contradiction} \\
        \midrule
        \multicolumn{2}{c}{$\cdots$} \\
        \bottomrule
        \end{tabular}
    }
    \caption{Examples of original NLI data and counterfactual data generated by AutoCAD. The identified rationales in the original data and generated spans in the counterfactual data are highlighted in colors along with the corresponding labels.}
    \label{tab:examples}
\end{table}

State-of-the-art NLU models have achieved impressive in-distribution performance, even surpassing humans on many benchmarks such as GLUE \cite{wang2018glue} and SuperGLUE \cite{wang2019superglue}.
However, these apparently powerful NLU models are known to suffer from shortcut learning, i.e., learning spurious features in datasets instead of actually solving the underlying tasks, thereby leading to unsatisfactory generalizability \cite{geirhos2020shortcut}. For example, in Natural Language Inference, a negation word can be a strong indicator of contradiction, and a high rate of word overlap between the premise and the hypothesis can be a strong indicator of entailment \cite{gururangan2018annotation, naik2018stress}. This phenomenon hinders the practical application of NLU models.

Among the recent attempts to mitigate shortcut learning in deep neural networks, Counterfactually Augmented Data (CAD) attracts increasing attention due to its simplicity and effectiveness \cite{kaushik2019learning}. Specifically, CAD are created through altering the ground-truth labels of original examples by manipulating causal features. By adding CAD to the original dataset, we can reduce spurious correlations between non-causal features and labels. The shortcut learning problem is hence mitigated from a data-centric perspective, yielding classifiers with better generalizability.

Different from the traditional label-preserving data augmentation methods, e.g., simply replacing random words with their synonyms, the creation of CAD is more challenging as it requires precisely disentangling spurious and causal features, as well as making proper changes to alter the label. To tackle this problem, many existing studies still rely on human efforts to create CAD \cite{kaushik2019learning,gardner2020evaluating} or label the generated CAD \cite{wu2021polyjuice}, which is costly and time-consuming. Moreover, manually-curated CAD may even exacerbate existing spurious patterns or introduce new ones due to the lack of diversity in annotators' edits \cite{huang2020counterfactually, joshi2021investigation}. While task-specific methods can be designed to automatically generate counterfactuals for sentiment analysis \cite{yang2021exploring} and open-domain QA\cite{paranjape2021retrieval}, the design and effectiveness of a fully automatic and task-agnostic CAD generator are still under-explored.

In this paper, we present AutoCAD, a fully automatic and task-agnostic CAD generation framework, which only takes the original NLU dataset as input and can generate diverse counterfactuals guided by any given target label. Our framework design completely eliminates the need for human efforts, task-specific designs, or parallel counterfactually augmented data for supervised training. Specifically, AutoCAD first leverages the gradients of a trained classifier to automatically identify rationales as spans to be intervened. Then, AutoCAD formulates counterfactual generation as a label-controlled text-infilling task with the help of a large-scale sequence-to-sequence language model. AutoCAD further introduces unlikelihood training \cite{welleck2019neural} to improve the controllability of counterfactual generation.

We study the effectiveness of AutoCAD on two widely-adopted NLU tasks: Natural Language Inference and Sentiment Analysis. We extend the original training data with counterfactuals automatically generated by AutoCAD and evaluate the generalizability of state-of-the-art NLU models on multiple out-of-domain and challenge benchmarks. Extensive experiments demonstrate that AutoCAD consistently and significantly improves the out-of-distribution performance across different tasks and different powerful pre-trained models. AutoCAD outperforms previous non-CAD data augmentation baselines by a large margin, especially on the challenging benchmarks where shortcut learning behavior is amplified. It also achieves comparable or even better results than previous state-of-the-art human-in-the-loop or task-specific CAD methods.

\section{Related Work}

\subsection{Mitigating Shortcut Learning of NLU Models}

There are two lines of research towards mitigating shortcut learning of NLU models: model-centric and data-centric.
Model-centric methods focus on reducing the reliance on spurious features during the training phase of NLU models.
\citet{clark2019don, he2019unlearn, mahabadi2019end} propose to ensemble a bias-only model with the main model based on the product-of-experts(PoE) framework to suppress it from focusing on the known dataset-specific bias. \citet{utama2020towards, du2021towards} further propose general methods to quantify the shortcut degree of each sample without the prior knowledge of bias and then debias NLU models through re-weighting or confidence regularization during training. Instead of incorporating additional modules or training objectives, data-centric methods focus on intrinsically reducing the spurious features in datasets. \citet{wu2022generating} proposes a filtering mechanism based on $z$-statistics \cite{gardner2021competency} for removing data samples that contribute to spurious features. And CAD also falls into this category.

\subsection{Counterfactually Augmented Data}

\citet{kaushik2019learning, gardner2020evaluating} employ human annotators to create CAD by manually rewriting the original datasets. However, manual rewrites are not only time-consuming and expensive but also may exacerbate existing spurious features or introduce new ones due to the lack of diversity in annotators' edits \cite{huang2020counterfactually, joshi2021investigation}. To alleviate this issue, \citet{wu2021polyjuice} propose \textsc{polyjuice}, a task-agnostic GPT2-based counterfactual generator, which is fine-tuned on parallel original-counterfactual pairs collected from multiple datasets to allow for control codes such as \textit{negation}, \textit{insert} and \textit{delete}. However, since \textsc{polyjuice} is an untargeted counterfactual generator, human annotators are still needed to label the generated counterfactuals. There are also some task-specific methods to automatically generate counterfactuals for sentiment analysis \cite{yang2021exploring} and open-domain QA\cite{paranjape2021retrieval}. Similar to our work, \citet{madaan2021generate,ross2020explaining} also aims to design a task-agnostic automatic counterfactual generator. Our work is distinguished from these mainly in that we not only evaluate the performance of the proposed generator in controllability and diversity but also conduct extensive experiments on multiple out-of-domain and challenge benchmarks to thoroughly investigate the efficacy of the automatically generated CAD for improving the generalizability of powerful pre-trained NLU models across different tasks.

\subsection{Controllable Text Generation}

Controllable text generation aims to generate texts aligning with the desired attribute and hence is an essential component of automatic CAD generation. The most straightforward and commonly used approach is to directly fine-tune a generative model with the concatenation of the text and the targeted attribute, which is also known as Class-Conditional Language Model~(CCLM) \cite{keskar2019ctrl}. Another line of work achieves attribute control during the decoding process without updating parameters of large-scale language models, which reduces computation costs \cite{dathathri2019plug,krause2020gedi,liu2021dexperts,yang2021fudge}. In general, CCLM performs better in text quality and inference speed but worse in controllability. We thus base our generator on CCLM and further introduce unlikelihood training to enhance its controllability. 
\section{Methodology}

\subsection{Task Definition}

Let $\{(x_i,y_i)\}$ be the training set of a classification task, where $x_i$ can either be a text or a text pair, and $y_i \in \mathcal{Y}$ is the corresponding label.

For each $(x_i,y_i)$ and a given target label $\hat{y_i} \ne y_i$, we define the counterfactual generation task as to generate one or more counterfactual example $\hat{x_i}$ that meet the following requirements.
1) \textbf{\textit{Label flipping}}: ${\hat{x_i}}$ should be in accord with the target label $\hat{y_i}$. 
2) \textbf{\textit{Fluent}}: $\hat{x_i}$ should be coherent and grammatically correct.
3) \textbf{\textit{Minimal change}}:

Ideally, $\hat{x_i}$ should only intervene in the rationales (causal features) to disentangle them from the spurious features. The requirement of \textit{minimal change} is positively correlated with \textit{label flipping} since it is less likely to flip the original label $y_i$ without touching its rationales.
4) \textbf{\textit{Diverse}}: the changes from $x_i$ to $\hat{x_i}$ should be diverse among the dataset, especially when giving the same target label $\hat{y_i}$. Otherwise, the generated counterfactuals may exacerbate existing spurious features or introduce new spurious features \cite{huang2020counterfactually, joshi2021investigation}.

\begin{figure*}[t]
  \centering
  \includegraphics[width=\linewidth]{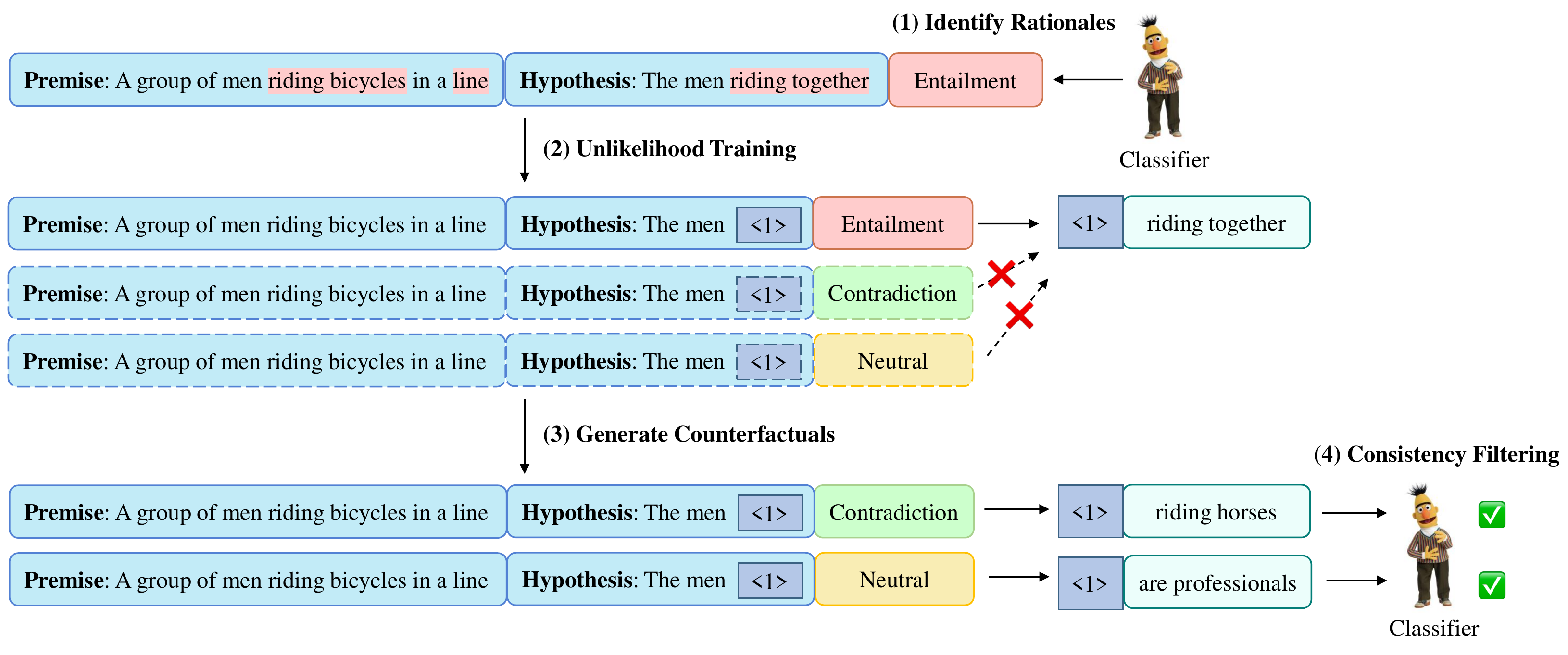}
  \caption{
    Overview of AutoCAD. The framework consists of four steps: (1) leverage a classifier to unsupervisedly identify rationales as spans to be intervened (masked); (2) train a controllable text-infilling generator enhanced by unlikelihood training; (3) generate counterfactuals according to a flipped target label; (4) post-select the generated samples based on the agreement of the target label and the predicted label of the classifier.
  }
  \label{fig:autocad}
\end{figure*}

To tackle the problem, AutoCAD first leverages a trained classifier to unsupervisedly identify rationales as spans to be intervened, eliminating the need for human efforts or task-specific designs. Then, AutoCAD formulates counterfactual generation as a label-controlled text-infilling task and introduces unlikelihood training \cite{welleck2019neural} to improve the controllability of counterfactual generation, eliminating the need for parallel CAD for supervised training. After generating $\hat{x_i}$, AutoCAD further uses the classifier to post-select the generated samples. Figure \ref{fig:autocad} shows the framework overview of AutoCAD.

\subsection{Identifying Rationales} \label{identify}

To meet the requirement of \textit{minimal change} and \textit{label flipping}, we need to select the spans to be intervened carefully. Ideally, changing the words exactly belonging to rationales will be the most effective approach. However, the golden rationales are unavailable without human efforts. Therefore, we adopt the existing task-agnostic post-hoc explanation methods to automatically identify rationales. 
There are two mainstream categories of post-hoc explanation methods: perturbation-based \cite{ribeiro2016should, li2016understanding} and gradient-based \cite{simonyan2013deep,li2015visualizing}. In this work, we implement AutoCAD with the gradient norm \cite{li2015visualizing}, and our framework can be easily adapted to any other rationale extraction method. 

Formally, given a classification model trained on the original dataset and a data sample $(x, y)$, where $x = (w_1, w_2, \cdots, w_m) = (t_1, t_2, \cdots, t_n)$ is the text input with $m$ words and $n$ tokens, and $y$ is the classification label, we first calculate the gradient of the model output with respect to the embedding $e_i$ of each token $t_i$ in the input $x$:

$$
g(t_i) = \nabla_{e_i} P(y|x)
$$
where $P(y|x)$ is the output logit for label $y$ given the input $x$. 

Then we obtain the saliency score $s_{t_i}$ of each token $t_i$ by taking the $l2$ norm of the gradient and re-normalizing it along all the input tokens:

\begin{align*}
    s_{t_i} = \frac{||g(t_i)||_2}{\sum_{j=1}^n||g(t_j)||_2}
\end{align*}

Considering that some tokenization algorithms adopted in widely used NLU Models, e.g., byte-level Byte-Pair-Encoding (BPE) in RoBERTa \cite{liu2019roberta}, would split a single word $w_i$ into $K$ sub-tokens $(t^1_i, t^2_i, \cdots, t^K_i)$, we further derive the word-level saliency score $s_{w_i}$ as the largest saliency score of its sub-tokens:

$$
s_{w_i} = \max_{k=1}^K s_{t^k_i}
$$

After obtaining the word-level saliency score, we select the top $\pi\%$ words with the largest $s_{w_i}$ values as rationales, where $\pi$ is a threshold hyperparameter. Note that the classifier we used is trained on the original dataset and thereby is also exposed to spurious features. In order to mitigate error propagation, we only select those samples for which the model predictions are correct. We assume the model is more likely to exploit the golden rationales in these samples than the rest.

\subsection{Unlikelihood Training for Label-Controlled Text Infilling} \label{generator}

In this section, we aim to train a generator that can modify the identified rationales in a diverse way and produce counterfactual data in accord with the new label. As we do not have parallel CAD for supervised training like \textsc{polyjuice}, we formulate counterfactual generation as a label-controlled text-infilling task, i.e., to generate more variations for the masked rationale spans in accord with a given flipped label. Moreover, this task formulation also ensures that all the identified non-rationales will remain unchanged in the generated counterfactuals. 

We base our generation model on an encoder-decoder model - T5 \cite{raffel2019exploring} particularly - for two reasons.
1) Controllability and diversity are both crucial for the effectiveness of CAD. As revealed by \citet{kumar2020aug}, texts generated by encoder-decoder models achieve a good balance between controllability and diversity compared with
encoder-only models like BERT \cite{devlin2018bert} or decoder-only models like GPT \cite{radford2018improving}.
2) T5 is pre-trained with the same text-infilling objective, which mitigates the gap between pre-training and fine-tuning. Thus we can take better advantage of knowledge in pre-trained language models for generating counterfactuals.

In order to realize label-controlled generation, the common training objective of CCLM is to minimize the negative log-likelihood loss $\mathcal{L}_{MLE}$ of reconstructing the text output conditioned on the given label. In AutoCAD, the loss is as follows:

$$
\mathcal{L}_{MLE} =-\sum_{t=1}^{|z|}\text{log}P(z_t|z_{<t}, x - z, y)
$$
where $z$ denotes the identified rationales, $x$ is the original text input, $x - z$ is the rationale-corrupted text input, and $y$ is the classification label.

However, using only the likelihood training objective may lead the model to over-focus on the text input $x - z$ and ignore the control label $y$. 

To alleviate this issue, we introduce unlikelihood training \cite{welleck2019neural} to suppress the model from always generating the original rationales when a perturbed label $\hat{y}$ is given. Formally, the unlikelihood loss $\mathcal{L}_{UL}$ is computed as follows:

$$
\mathcal{L}_{UL} = -\sum_{\hat{y}\in\mathcal{Y}, \hat{y}\neq y}\sum_{t=1}^{|z|}\text{log}(1-P(z_t|z_{<t}, x - z, \hat{y}))
$$

Then we derive the overall loss function as follows:

$$
\mathcal{L} = \mathcal{L}_{MLE} + \alpha * \mathcal{L}_{UL}
$$
where the coefficient $\alpha$ is a hyperparameter to control the strength of the unlikelihood loss.

Intuitively, minimizing the unlikelihood loss $\mathcal{L}_{UL}$ improves the model's sensitivity to the control label and thus results in better controllability on generation. However, we note that the token $z_t$ to be penalized needs to be carefully selected. As non-rationales have no strong association with any control label $y$, penalizing them may accidentally introduce noise to the generator and degrade its performance, which is validated in our experiment results in Section \ref{subsection_result}.

\subsection{Consistency Filtering}

Since we aim at a task-agnostic and fully-automatic counterfactual generation framework, the controllability of the generator could be insufficient when applied to some challenging NLU tasks such as Natural Language Inference. Therefore, we further leverage the classification model used in Section~\ref{identify} to post-select the generated output. Specifically, we deem a generated counterfactual as valid if its target label is consistent with the prediction of the classifier. After consistency filtering, we finally obtain the counterfactually augmented data.

\section{Experiments}

\subsection{Benchmark Tasks}

We evaluate our method on two widely-adopted NLU tasks, Natural Language Inference (NLI) and Sentiment Analysis (SA). We select the SST-2 dataset which contains 8,544 examples \cite{socher2013recursive} and a subset from the SNLI dataset which contains 20,000 examples \cite{bowman2015large} as the original in-domain dataset. And since we focus on measuring the generalizability of the model after data augmentation, we further conduct extensive evaluations on multiple out-of-domain and challenge benchmarks for each task. For more details about the evaluation benchmarks, please refer to Appendix \ref{section: eval_dataset}.

\subsection{Baselines}

We compare AutoCAD with the following baselines. For all the automatic data augmentation methods, we augment each sample once unless otherwise specified. 

\paragraph{Synonym Replacement} A proportion of $r$ words are chosen from the sentence (excluding the stopwords), and each chosen word is independently replaced with its synonyms based on WordNet \cite{zhang2015character}. The $r$ is set to $30\%$ in our experiment.

\paragraph{Back Translation} The sentence is first translated to another language, and then back-translated to the source language \cite{sennrich2015improving}. 

\paragraph{BERT-MLM} A proportion of $r$ words are chosen from the sentence excluding the stopwords, and each chosen word is independently replaced with the top $k$ words predicted by a pre-trained BERT model \cite{jiao2019tinybert}. The $r$ is set to $30\%$ and $k$ is set to 100 in our experiment.

\paragraph{Sentiment-CAD} Sentiment-CAD \cite{yang2021exploring} is the previous state-of-the-art method to automatically generate counterfactuals for sentiment analysis. Sentiment-CAD uses an external sentiment word dictionary and a sentiment classifier to identify rationales. For each word in the identified rationales, it either removes the word (usually a negation word such as ``not'' and ``no'') or generates an alternative word with an opposing sentiment polarity according to the sentiment word dictionary using BERT-MLM. However, there are two drawbacks of this method. First, the reliance on the external word dictionary makes Sentiment-CAD a task-specific method that may face the challenge when handling complex data without explicit sentiment words. Second, the diversity of generated data is limited as the encoder-only model used in Sentiment-CAD can only generate fix-length alternatives that are further constrained by the pre-defined dictionary.

\paragraph{Human-CAD} The Human-CAD dataset \cite{kaushik2019learning} is created by employing human annotators to rewrite a subset of the SNLI dataset. In our experiment, we ensure that our training dataset includes all the samples in the subset.

\smallskip \noindent We also compare AutoCAD with other methods that are not specially designed for CAD (see Appendix~\ref{sec:morebaselines}).

\subsection{Implementation Details}

We adopt the state-of-the-art pre-trained NLU model RoBERTa\textsubscript{\tiny{LARGE}} \cite{liu2019roberta} as the classifier to identify rationales and do consistency filtering, and T5\textsubscript{\tiny{LARGE}} \cite{raffel2019exploring} as the generator. For training the classifier, we set the batch size to 32, the initial learning rate of the AdamW optimizer to 1e-5, and the maximum training epoch to 20. We select the best checkpoint based on the accuracy of the in-domain validation set. For training the generator, we set the batch size to 8, the initial learning rate of AdamW to 1e-5, and the maximum training epoch to 10 with an early stopping mechanism. We set $\alpha=1$ for Natural Language Inference, and $\alpha=0.3$ for Sentiment Analysis. We select the best checkpoint based on the perplexity of the validation set. We generate counterfactuals using nucleus sampling \cite{holtzman2019curious} with $p=0.9$ and $\text{temperature}=0.7$.

\subsection{Main Results}\label{subsection_result}

The results on NLI and SA are shown in Table~\ref{tab:auto_main}. It can be seen that methods based on CAD generally outperform non-CAD methods. In particular, on NLI, which is a more challenging NLU task, we observe that all the non-CAD baselines result in negative in-distribution and out-of-distribution performance, except in only a few cases where a slight improvement can be achieved.
In contrast, AutoCAD consistently and significantly improves out-of-distribution performance while maintaining or slightly improving the in-distribution performance across different tasks and different pre-trained models. Especially that AutoCAD achieves impressive results on the challenge sets where shortcut learning behavior is amplified, demonstrating its effectiveness for mitigating shortcut learning. Furthermore, it also achieves comparable or even better performance than Human-CAD and Sentiment-CAD, while eliminating the need for any human effort or task-specific design.

\begin{table*}[t]
    \centering
    \resizebox{\textwidth}{!}
    {
        \begin{tabular}{lcccccccc|cccccc}
        \toprule
        \multicolumn{9}{c}{\textbf{Natural Language Inference}} & \multicolumn{6}{c}{\textbf{Sentiment Analysis}}\\
        \midrule
        \multirow{3}{*}{\textbf{Method}} & \multicolumn{1}{c}{\textbf{In-Domain}} & \multicolumn{2}{c}{\textbf{Out-of-Domain}} & \multicolumn{4}{c}{\textbf{Challenge}} & \multirow{3}{*}{\textbf{Avg. }} & \multicolumn{1}{c}{\textbf{In-Domain}} & \multicolumn{2}{c}{\textbf{Out-of-Domain}} & \multicolumn{2}{c}{\textbf{Challenge}} & \multirow{3}{*}{\textbf{Avg. }} \\
        \cmidrule(lr){2-2} \cmidrule(lr){3-4} \cmidrule(lr){5-8} \cmidrule(lr){10-10} \cmidrule(lr){11-12} \cmidrule(lr){13-14} 
        & SNLI & MNLI-m & MNLI-mm & Human-CAD
        & Diagnostic & Stress & Break & & SST-2 & IMDb & Yelp & Human-CAD & Contrast\\
        \midrule
        \midrule
        \multicolumn{15}{c}{\it{BERT\textsubscript{\tiny{BASE}}}}\\
        \midrule
        Original         & 84.84 & 63.02 & 63.84 & 61.25 & 50.27 & 54.55 & 69.32 & 60.38 & 88.42 & 86.68 & 88.95 & 87.50 & 82.58 & 86.43\\ 
        Synonym Rep. & 84.77 & 64.06 & 64.61 & 61.44 & 49.91 & 57.26 & 68.40 & 60.95 & 87.42 & 86.07 & 90.82 & 87.90 & 83.40 & 87.05\\ 
        Back Trans.  & 84.86 & 63.89 & 64.04 & 61.25 & 49.73 & 56.91 & 66.35 & 60.36 & 87.60 & \textbf{89.14} & 89.27 & 87.50 & 83.81 & 87.43\\ 
        BERT-MLM     & 83.92 & 62.78 & 63.88 & 57.18 & 49.73 & 56.30 & 66.25 & 59.35 & 87.06 & 84.84 & 88.53 & 80.12 & 77.25 & 82.69\\ 
        \midrule
        Human-CAD    & 85.75 & 66.26 & 66.14 & 70.87 & 51.72 & 57.74 & 79.38 & 65.35 & - & - & - & - & -\\ 
        Sentiment-CAD & - & - & - & - & - & - & - & - & 87.73 & 88.93 & 89.73 & 90.16 & 87.09 & 88.98 \\
        AutoCAD      & \textbf{87.25}& \textbf{69.67} & \textbf{70.27} & \textbf{71.43} & \textbf{54.26} & \textbf{59.13} & \textbf{89.59} & \textbf{68.38} & 88.19 & 88.52 & \textbf{90.94} & \textbf{91.80} & \textbf{88.73} & \textbf{90.00}\\ 
        \midrule
        \midrule
        \multicolumn{15}{c}{\it{BERT\textsubscript{\tiny{LARGE}}}}\\
        \midrule
        Original      & 86.15 & 69.32 & 70.15 & 66.75 & 53.80 & 60.92 & 83.58 & 67.42 & 87.38 & 85.45 & 87.88  & 88.52 & 83.40 & 86.31\\ 
        Synonym Rep. & 86.77 & 70.85 & 71.70 & 66.87 & 54.53 & 64.27 & 80.69 & 68.15 & 88.00 & \textbf{87.50} & 88.57  & \textbf{92.01} & 84.43 & 88.13\\ 
        Back Trans.  & 86.40 & 68.61 & 68.71 & 64.75 & 54.53 & 60.31 & 76.41 & 65.55 & 87.38 & 80.53 & 83.38  & 80.12 & 74.59 & 79.66\\ 
        BERT-MLM     & 84.38 & 65.61 & 66.74 & 58.50 & 50.72 & 58.69 & 69.18 & 61.57 & 88.05 & 87.30 & 88.54  & 84.84 & 79.51 & 85.05\\ 
        \midrule
        Human-CAD    & 86.88 & 70.46 & 69.36 & \textbf{73.87} & 53.89 & 61.78 & 90.37 & 69.96 & - & - & - & - & -\\ 
        Sentimen-CAD  & - & - & - & - & - & - & - & - & 88.87 & 85.04 & 88.56 & 88.73 & 85.66 & 87.00\\
        AutoCAD      & \textbf{87.98} & \textbf{74.50} & \textbf{75.05} & 73.75 & \textbf{55.98} & \textbf{65.04} & \textbf{90.53} & \textbf{72.22} & 87.78 & 86.27 & \textbf{89.94} & 90.98 & \textbf{86.68} & \textbf{89.04}\\ 
        \midrule
        \midrule
        \multicolumn{15}{c}{\it{RoBERTa\textsubscript{\tiny{BASE}}}}\\
        \midrule
        Original           & 88.02 & 75.07 & 76.07 & 68.31 & 55.07 & \textbf{67.09} & 91.22 & 72.14 & 91.09 & 85.66 & 91.16 & 85.45 & 81.56 & 85.96\\ 
        Synonym Rep.   & 87.61 & 73.91 & 75.23 & 67.44 & 55.34 & 65.43 & 84.79 & 70.32 & 89.68 & 87.70 & 93.26 & 90.16 & 87.70 & 89.71\\ 
        Back Trans.    & 87.86 & 74.11 & 74.75 & 67.12 & 54.34 & 65.02 & 80.10 & 69.28 & 91.18 & 87.50 & 92.22  & 88.52 & 84.84 & 88.27\\ 
        BERT-MLM       & 87.13 & 73.25 & 74.08 & 66.87 & 54.53 & 64.83 & 89.59 & 70.53 & 89.55 & 87.09 & 91.46  & 85.86 & 80.12 & 86.13\\ 
        \midrule
        Human-CAD      & 87.42 & 75.85 & 76.01 & \textbf{75.56} & 56.52 & 64.86 & 90.96 & 73.29  & - & - & - & - & -\\ 
        Sentiment-CAD  & - & - & - & - & - & - & - & - & 89.95 & \textbf{89.95} & \textbf{93.41} & \textbf{94.26} & \textbf{91.19} & \textbf{92.21}\\
        AutoCAD        & 88.11 & \textbf{76.32} & \textbf{77.09} & 74.56 & \textbf{56.88} & 65.17 & \textbf{91.77} & \textbf{73.63}  & 90.81 & 88.52 & 92.47 & 92.62 & 88.52 & 90.53\\ 
        \midrule
        \midrule
        \multicolumn{15}{c}{\it{RoBERTa\textsubscript{\tiny{LARGE}}}}\\
        \midrule
        Original         & 89.42 & 80.13 & 80.29 & 73.56 & 58.24 & 67.82 & 90.36 & 75.07  & 91.58 & 88.73 & 94.84 & 88.93 & 87.70 & 90.05\\ 
        Synonym Rep. & 89.58 & 80.83 & 81.77 & 72.62 & 56.70 & 72.33 & 91.42 & 75.95 & 90.41 & 89.55 & 94.57 & 89.55 & 90.37 & 91.01\\ 
        Back Trans.  & 88.92 & 78.75 & 78.95 & 72.37 & 59.51 & 67.38 & 90.54 & 74.58 & 89.68 & 91.60 & 95.49 & 92.62 & 91.39 & 92.78\\ 
        BERT-MLM     & 89.02 & 76.94 & 77.29 & 67.12 & 55.71 & 68.91 & 86.81 & 72.13 & 90.86 & 92.21 & \textbf{96.16} & 88.31 & 86.48 & 90.79\\ 
        \midrule
        Human-CAD    & \textbf{90.23} & 81.90 & \textbf{82.07} & \textbf{78.50} & 60.14 & 72.06 & \textbf{93.97} & 78.11  & - & - & - & - & - \\ 
        Sentiment-CAD & - & - & - & - & - & - & - & -  & 89.82 & 90.34 & 93.60 & 94.67 & 91.80 & 92.35\\
        AutoCAD      & 89.63 & \textbf{82.25} & 81.89 & 76.25 & \textbf{61.32} & \textbf{74.14} & 93.93 & \textbf{78.30} & 90.50 & \textbf{92.83} & 95.60 & \textbf{94.48} & \textbf{93.03} & \textbf{93.99}\\ 
        \bottomrule
        \end{tabular}
    }
    \caption{Model Accuracy on different NLU tasks with different data augmentation methods. The best results are highlighted in \textbf{bold}. The average scores take into account the scores on all the out-of-domain and challenge test sets. }
    \label{tab:auto_main}
\end{table*}

\subsection{Validity of the Identified Rationales}

\begin{figure}[htbp]
  \centering
  \includegraphics[width=\linewidth]{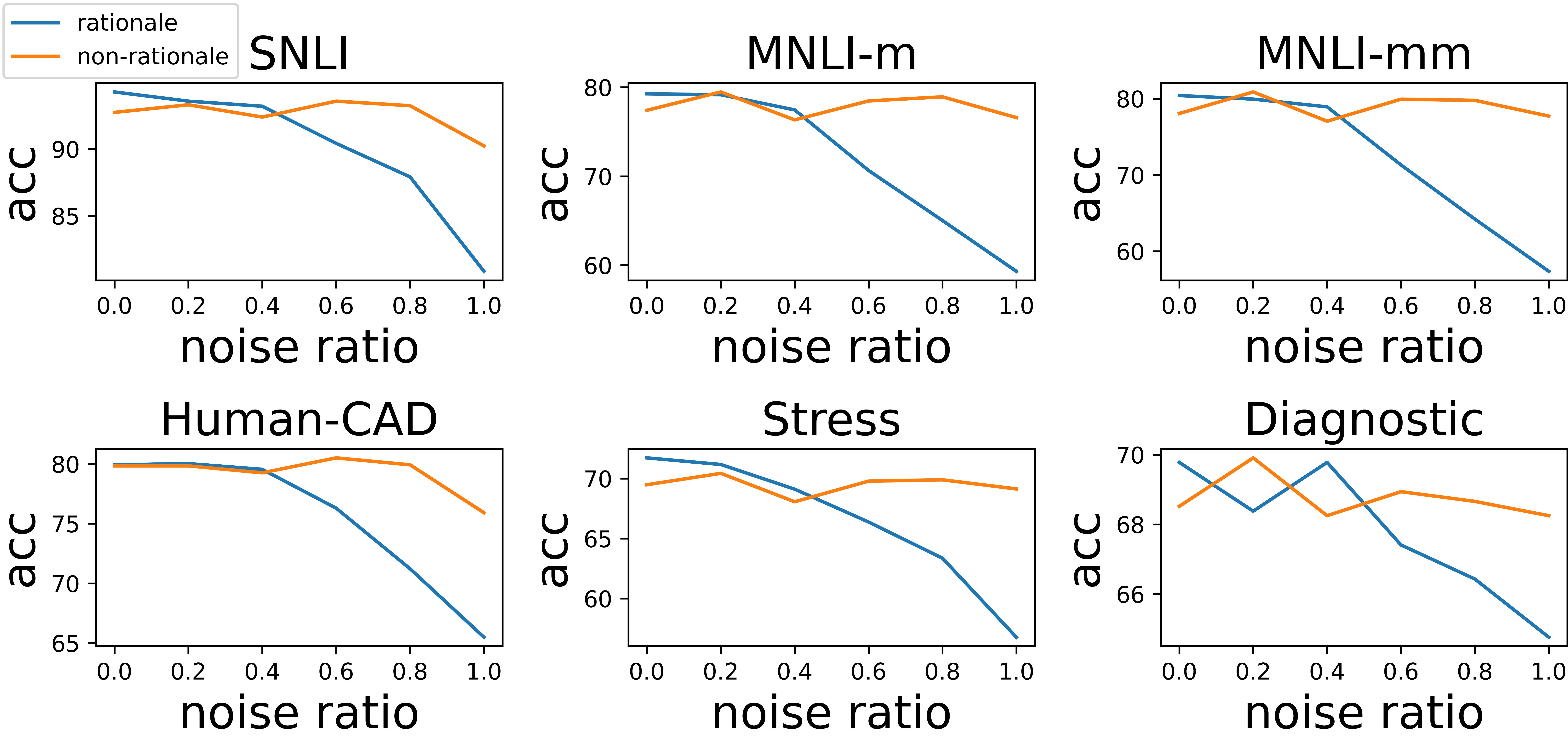}
  \caption{
    Changes in accuracy as we add noise to the unsupervisedly identified rationales or non-rationales. 
  }
  \label{fig:noise_exp}
\end{figure}

We first measure the alignment of model-identified rationales with human rationales from e-SNLI \cite{camburu2018snli, deyoung2019eraser}. The token-level macro-F1 score is 0.46.
Following \citet{kaushik2020explaining}, we further conduct experiments to verify the validity of the identified rationales. We randomly mask $r\%$ tokens in the identified rationales and non-rationales respectively, and observe performance changes of the NLU model trained on these two differently noised datasets. If the identified rationales can represent causal features, masking these rationales is expected to result in worse model performance than masking non-rationales. Furthermore, the difference should get amplified as the noising ratio $r\%$ increases. We conduct experiments on NLI using BERT\textsubscript{\tiny{BASE}}. As suggested by \citet{kaushik2020explaining}, we select only those premise-hypothesis pairs altogether with more than 9 tokens marked as rationales, eliminating the length imbalance between rationales and non-rationales. 
As shown in Figure~\ref{fig:noise_exp}, we observe a significantly sharper decrease in accuracy when adding noise to the identified rationales in all the test datasets, covering in-domain, out-of-domain, and challenge settings. The results demonstrate the internal validity of the identified rationales.

\subsection{Ablation Study}

To further investigate the influence of each component in AutoCAD, we run an ablation study on NLI and report the metrics of generation quality, including controllability measured by $label\, flipping\, rate$, and diversity by $distinct\textit{-}n$ \cite{madaan2021generate}. From the results in Table~\ref{tab:ablation}, it can be seen that combining rationale-masking strategy and unlikelihood training achieves the best performance on controllability and diversity. Moreover, we draw the following conclusions.
1) Masking rationales instead of random spans can effectively improve the label flipping rate, in both the training and generation phases and in different loss settings.
2) While the controllability of the generator improves after fine-tuning with standard MLE objective, combining with unlikelihood training can further boost the label flipping rate from 42.17\% to 68.56\%.
3) Unlikelihood training significantly improves the diversity of generation, as the generator is forced to generate under the guidance of the target label other than just generating the words seen in the original example.
4) Masking rationales can fully exploit the benefits of unlikelihood training, resulting in a substantial controllability improvement from 47.87\% to 68.56\%. In fact, when using random mask in unlikelihood training, we observe that the unlikelihood loss $\mathcal{L}_{UL}$ will gradually conflict with the likelihood loss $\mathcal{L}_{MLE}$.

\begin{table}[tbp]
    \centering
    \resizebox{\linewidth}{!}
    {
        \begin{tabular}{lcccccc}
        \toprule
        Variants (train loss) & Mask$_{train}$ & Mask$_{gen}$ & FR & Distinct-3/4  \\
        \midrule
        AutoCAD$_{notrain}$ & N/A & random & 23.65 & 0.25/0.94 \\
        \midrule
        AutoCAD$_{MLE}$     & random & random & 32.17 & 0.33/1.18\\
        AutoCAD$_{MLE}$     & random & rationales & 34.09  & 0.23/0.83\\
        AutoCAD$_{MLE}$     & rationales & rationales & 42.17 & 0.27/0.94\\
        \midrule
        AutoCAD$_{MLE+UL}$ & random & rationales & 47.87 & 0.39/\textbf{1.61}\\
        AutoCAD$_{MLE+UL}$ & rationales & rationales & \textbf{68.56} &  \textbf{0.40}/1.48\\
        \bottomrule
        \end{tabular}
    }
    \caption{Ablation study on the effect of rationales and unlikelihood training on generator's performance, conducted on SNLI. FR means label flipping rate. (Refer to Appendix \ref{section: metrics} for more details about the metrics.)}
    \label{tab:ablation}
\end{table}

\subsection{Analysis of $\alpha$ in Unlikelihood Training}

\begin{figure}[htbp]
  \centering
  \includegraphics[width=\linewidth]{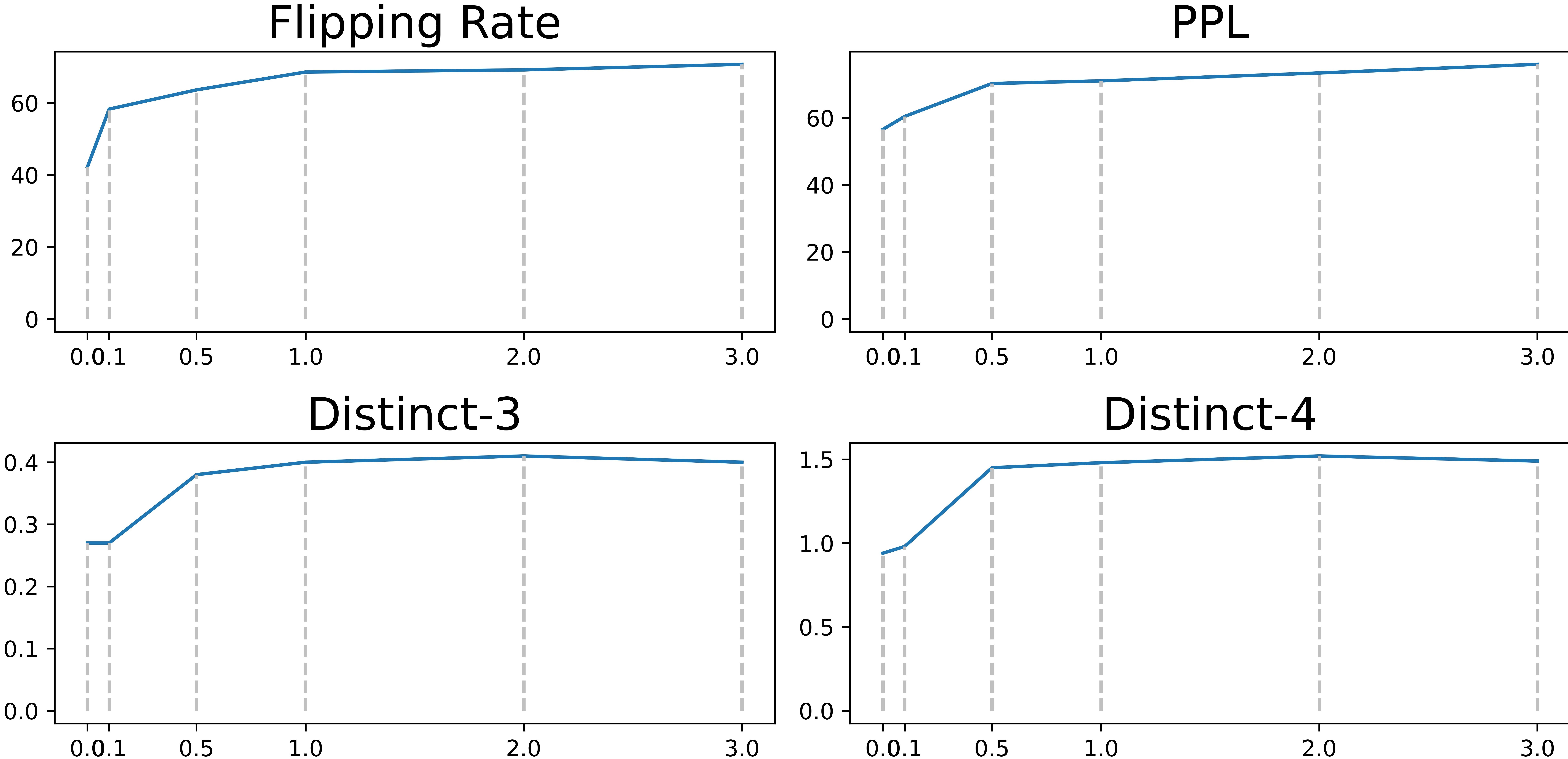}
  \caption{
    The effect of the coefficient $\alpha$ on the performance of the generator trained on SNLI.
  }
  \label{fig:coefficient}
\end{figure}

We investigate the effect of the coefficient $\alpha$ on the generator's performance on the NLI task. As shown in Figure \ref{fig:coefficient}, as $\alpha$ increases, there is a significant improvement in label flipping rate and diversity, with a modest increase in perplexity. This trend slows down after the coefficient $\alpha$ exceeds 1.0.

\subsection{Comparing AutoCAD with FlipDA}

\begin{table*}[htbp]
    \centering
    \small
    {
      \begin{tabular}{lccccccccc}
        \toprule
        \multirow{3}{*}{\textbf{Method}(Augmenting Times)} & \multicolumn{1}{c}{\textbf{In-Domain}} & \multicolumn{2}{c}{\textbf{Out-of-Domain}} & \multicolumn{4}{c}{\textbf{Challenge}} & \multirow{3}{*}{\textbf{Avg. (\%   )}} \\
        \cmidrule(lr){2-2} \cmidrule(lr){3-4} \cmidrule(lr){5-8} 
        & SNLI & MNLI-m & MNLI-mm & CAD
        & Diagnostic & Stress & Break \\
        \midrule
        Original         & 84.84 & 63.02 & 63.84 & 61.25 & 50.27 & 54.55 & 69.32 & 60.38\\
        \midrule
        FlipDA  (10)      & 86.07 & 68.81 & 69.27 & 67.19 & 53.17 & 57.73 & 82.49 & 66.44\\
        AutoCAD (1)      & 87.25 & 69.67 & 70.27 & 71.43 & 54.26 & 59.13 & 89.59 & 68.38\\
        AutoCAD (10)     & \textbf{87.61} & \textbf{72.35} & \textbf{72.96} & \textbf{72.31} & \textbf{55.71} & \textbf{60.14} & \textbf{92.19} & \textbf{70.94}\\
        \bottomrule
        \end{tabular}
    }
    \caption{Comparison between AutoCAD and FlipDA, conducted on BERT\textsubscript{\tiny{BASE}}. (1) and (10) after each method mean the number of augmenting times. CAD generated by AutoCAD are more sample efficient than those by FlipDA.}
    \label{tab:size_quality}
\end{table*}

\begin{figure*}[ht]
  \centering
  \includegraphics[width=\linewidth]{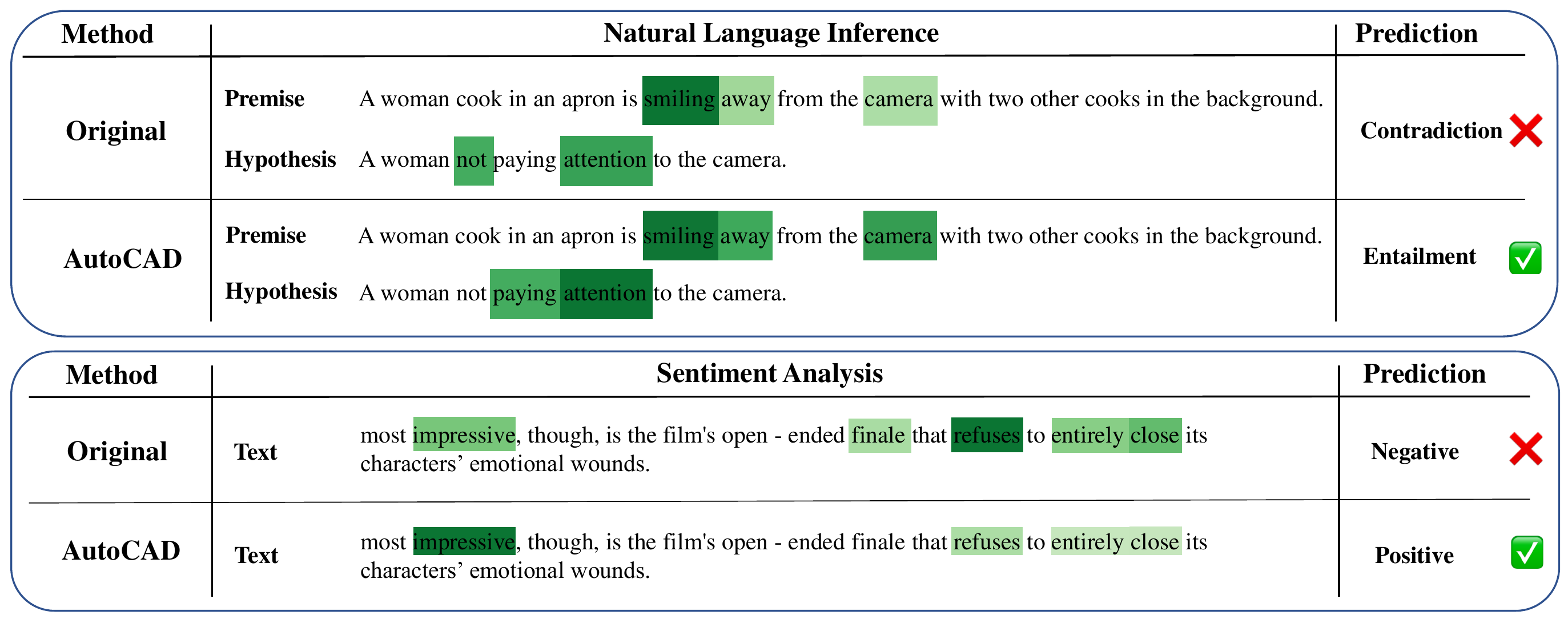}
  \caption{
    Examples of the effect of AutoCAD on downstream NLU tasks. We observe the prediction changes of a BERT\textsubscript{\tiny{BASE}} model trained with or without data augmentation by AutoCAD. The highlighted words are the top-$5$ rationales identified by the method in Section \ref{identify}, indicating the importance of each word to the model decisions.
  }
  \label{fig:case_classifier}
\end{figure*}

We note that AutoCAD$_{notrain}$ is similar to FlipDA \cite{zhou2021flipda}, a concurrent work that focuses on improving in-domain performance in the low-resource scenario only and gets substantial improvement on various NLU tasks. While both AutoCAD$_{notrain}$ and FlipDA randomly select spans to be intervened and leverage the vanilla T5 model to fill in the blank, FlipDA has a more specific design on prompt engineering and post-selection strategy. Therefore, we adopt the same setting from their paper to fully exploit its performance: (1) we use ``\textit{\{premise\} ? Yes/No/Maybe . \{hypothesis\}}'' as the prompt. (2) we augment $N=10$ times for each example in the original dataset. (3) we select the generated counterfactual with the largest probabilities for the target label. 

As shown in Table~\ref{tab:size_quality}, AutoCAD consistently and substantially outperforms FlipDA while using fewer counterfactuals after augmenting only once. Moreover, augmenting $N=10$ times with AutoCAD will further boost the performance of NLU models. The results indicate that the counterfactuals generated by AutoCAD are more informative and sample efficient compared with FlipDA. We conjecture that the random mask strategy and the weaker controllability of FlipDA may lead to more label-preserving adversarial examples, and thereby introduce more label noise.

\subsection{Case Study}

We present two cases from our NLU benchmark tasks in Figure~\ref{fig:case_classifier}. We can observe that the NLU model trained on the original data over-relies on words like ``not'' and ``refuses'', leading to wrong predictions. In contrast, counterfactually augmented data generated by AutoCAD effectively mitigates this phenomenon and successfully corrects the model predictions. We also present multiple generation examples in Appendix~\ref{sec:examples} to demonstrate that AutoCAD can generate diverse counterfactuals across different tasks.

\subsection{Discussion}

By disentangling instance-level causal and non-causal features, CAD could implicitly mitigate unknown statistical spurious features in dataset-level \cite{kaushik2019learning, kaushik2020explaining}. The assumption here is different from that the prior of spurious features should be known and characterized with researchers' task-specific insights \cite{clark2019don, he2019unlearn, mahabadi2019end}. The latter has several limitations: 1) Datasets are likely to contain unknown spurious features which are hard to define. 2) The spurious features are dataset-specific.

\section{Conclusion}
We propose AutoCAD, a fully automatic and task-agnostic counterfactually augmented data generation framework. AutoCAD combines effective rationale extraction methods and a controllable generative model enhanced by unlikelihood training, which can generate diverse counterfactuals. Extensive experiments on multiple out-of-domain and challenge benchmarks demonstrate that AutoCAD consistently and significantly improves the out-of-distribution performance of powerful pre-trained NLU models. More importantly, AutoCAD achieves comparable or even better performance than previous state-of-the-art human-in-the-loop or task-specific methods. We believe this work has broad interests in various NLU tasks. 

\section{Limitations}
Despite the effectiveness of the gradient norm for rationale extraction in AutoCAD, we have not further explored more advanced methods such as LIME \cite{ribeiro2016should}, SHAP \cite{lundberg2017unified}, and L2E \cite{situ2021learning}. Another limitation of our work is that AutoCAD is still a pipeline framework. Some recent studies attempt to jointly optimize a rationale extractor and a classifier in an end-to-end fashion \cite{lei2016rationalizing, chang2020invariant, paranjape2020information, yu2021understanding}. We will extend AutoCAD to an end-to-end framework by jointly training a rationale extractor and a counterfactual generator.

\section*{Acknowledgement}
This work was supported by the National Science Foundation for Distinguished Young Scholars (with No. 62125604) and the NSFC projects (Key project with No. 61936010 and regular project with No. 61876096). This work was also supported by the Guoqiang Institute of Tsinghua University, with Grant No. 2019GQG1 and 2020GQG0005, and sponsored by Tsinghua-Toyota Joint Research Fund. 
We would also like to thank the anonymous reviewers for their invaluable suggestions and feedback.

\bibliography{anthology,custom}
\bibliographystyle{acl_natbib}

\appendix

\section{Appendix}
\label{sec:appendix}

\subsection{Metrics} \label{section: metrics}

\paragraph{Distinct-N} It is computed as the number of distinct n-grams divided by the number of all n-grams in generated data \cite{li2015diversity}.

\paragraph{Label Flipping Rate (FR)} Ideally, it is computed as the proportion of the generated counterfactual data whose target label is consistent with its golden label. Since the golden label is unavailable without human efforts, we use the prediction of the classifier instead. Formally,
$$
\text{FR} = \frac{\sum_{j=1}^N \mathbbm{1}[\hat{y} = \arg\max_{y\in \mathcal{Y}} P(y|\hat{x})]}{N}
$$
where N is the size of all the generated counterfactuals and $\mathbbm{1}[\cdot]$ is the indicator function.

\subsection{Details of Evaluation Benchmarks} \label{section: eval_dataset}

\subsubsection{Natural Language Inference}
\begin{itemize}
    \item \textbf{MNLI-m and MNLI-mm} \cite{williams2017broad}: 
    The matched and the mismatched test set of the Multi-Genre NLI dataset (MultiNLI) differ in text domains. MultiNLI is derived from ten different text genres of written and spoken English, and is more challenging compared with the Standford NLI dataset (SNLI) \cite{bowman2015large} which is derived from only one domain. 
    
    \item \textbf{Human-CAD} \cite{kaushik2019learning}: The Human-CAD dataset for NLI is a manually-curated counterfactually augmented dataset created by employing human annotators to rewrite a subset of the SNLI dataset.
    
    \item \textbf{Diagnostic} \cite{wang2018glue}: The Diagnostic dataset is a manually-curated test set for evaluating the model's ability on several important linguistic phenomena, such as lexical semantics and logic.
    
    \item \textbf{Stress} \cite{naik2018stress}: The Stress test set reveals a model's ability to reason about antonyms and numbers, reliance on spurious lexical features, and robustness to random perturbations. It is constructed based on error analysis and creating adversarial examples from MultiNLI.
    
    \item \textbf{Break} \cite{glockner2018breaking}: The Breaking NLI dataset is an adversarial test set for evaluating the ability of lexical inferences. For each premise in the SNLI dataset, several hypotheses are generated by replacing a single word in the premise and manually verified by crowd-sourced workers. 
\end{itemize}

\subsubsection{Sentiment Analysis}
\begin{itemize}
    \item \textbf{IMDb} \cite{maas2011learning}: The Internet Movie Database (IMDb) dataset is a collection of movie reviews.
    
    \item \textbf{Yelp} \cite{asghar2016yelp}: The Yelp Dataset is a collection of user reviews for businesses, products, and services.
    
    \item \textbf{Human-CAD} \cite{kaushik2019learning}: The Human-CAD dataset for sentiment analysis is a counterfactually augmented dataset created by employing human annotators to rewrite a subset of the IMDb dataset.
    
    \item \textbf{Contrast} \cite{gardner2020evaluating}: Similar to Human-CAD, Contrast is also a counterfactually augmented dataset created by manually rewriting a subset of the IMDb dataset.
\end{itemize}

\subsection{Experimental Details}

\subsubsection{Data Preprocessing}

\paragraph{\textbf{SNLI}} We extract the training set and validation set from the official split of SNLI while balancing the label distribution.

\paragraph{\textbf{SST-2}} We use the official split of SST-2 without further preprocessing.

\begin{table}[h]
    \centering
    \begin{tabular}{lcc}
    \toprule
    Dataset & Train & Dev \\
    \midrule
    SNLI & 20,000 & 2,400 \\
    SST-2 & 8,544 & 1,101 \\
    \bottomrule
    \end{tabular}
    \caption{Data Statistics.}
    \label{tab:dataset_stat}
\end{table}

\subsubsection{Training Details}

We provide more details about the training settings of our experiment.
Our codes are implemented based on Huggingface's Transformers \cite{wolf2020transformers}. Table~\ref{tab:parameter} shows the number of parameters for the models we used in our experiment. All experiments are carried out on a single V100 GPU (32GB). Each experiment can be completed in less than 10 hours. We use manual search to select the best hyperparameters, and the search space is presented in Table~\ref{tab:hyper}. Our model selection criterion is validation accuracy for the classifier and validation perplexity for the generator.

\begin{table}[h]
    \centering
    \begin{tabular}{lc}
    \toprule
    \textbf{Model} & \textbf{Number of Parameters} \\
    \midrule
    BERT\textsubscript{\tiny{BASE}} & 110M\\
    BERT\textsubscript{\tiny{LARGE}} & 340M\\
    RoBERTa\textsubscript{\tiny{BASE}} & 125M\\
    RoBERTa\textsubscript{\tiny{LARGE}} & 355M\\
    \midrule
    T5\textsubscript{\tiny{LARGE}} & 770M\\
    \bottomrule
    \end{tabular}
    \caption{Number of parameters.}
    \label{tab:parameter}
\end{table}

\begin{table}[h]
    \centering
    \small
    \begin{tabular}{lc}
    \toprule
    \textbf{Hyperparameter} & \textbf{Search Space} \\
    \midrule
    \midrule
    \multicolumn{2}{c}{\textbf{\textit{classifier}}}\\
    \midrule
    Learning Rate & \textit{choice}[1e-5, 5e-5]\\
    Training Epoch & \textit{choice}[5, 10, 20]\\
    Sequence Length & \textit{choice}[64, 128, 350]\\
    Optimizer & AdamW \\
    Epsilon (for AdamW) & 1e-8 \\
    Weight Decay & 1e-1 \\
    \midrule
    \midrule
    \multicolumn{2}{c}{\textbf{\textit{generator}}}\\
    \midrule
    Learning Rate & \textit{choice}[1e-3, 1e-4, 1e-5]\\
    Training Epoch & \textit{choice}[5, 10, 20]\\
    Sequence Length & \textit{choice}[128, 350]\\
    Optimizer & \textit{choice}[AdamW, Adafactor] \\
    Epsilon (for AdamW) & 1e-8 \\
    Weight Decay & 1e-1 \\
    Warmup Ratio & \textit{choice}[0, 0.01] \\
    \bottomrule
    \end{tabular}
    \caption{Number of parameters.}
    \label{tab:hyper}
\end{table}

\subsection{Comparing AutoCAD with More Baselines} \label{sec:morebaselines}

To our knowledge, there are limited automatic label-flipping baselines other than Sentiment-CAD and FlipDA, which are already presented in Table \ref{tab:auto_main} and Table \ref{tab:size_quality}. In this section, we further compare AutoCAD with three baselines, i.e., IRM \cite{dranker2021irm}, C2L \cite{choi2022c2l} and sentiment style-transfer \cite{shen2017style}.  We conduct experiments using BERT$_\text{BASE}$. Experiment results are shown in Table \ref{tab:morebaseline_nli} and Table \ref{tab:morebaseline_sst}. 
AutoCAD consistently and significantly outperforms all the three baselines. 
For IRM, we run the hypothesis-only setting from the official implementation. In line with \citet{dranker2021irm}, IRM shows a large variance and does not work in natural datasets. For C2L, we empirically choose the best $\lambda$ between [0.1,1.0] and find the gain is slight. In line with Sentiment-CAD, we find that the sentiment-flipped data generated by style-transferring degrades the performance. 

\begin{table}[h]
    \centering
    \begin{tabular}{lcc}
    \toprule
    Method & Avg. \\
    \midrule
    Original & 60.18$_{\pm1.9}$ \\
    IRM \cite{dranker2021irm} & 44.62$_{\pm5.1}$ \\
    C2L \cite{choi2022c2l} & 60.51$_{\pm1.1}$\\
    \midrule
    AutoCAD & \textbf{68.31$_{\pm0.6}$}\\
    \bottomrule
    \end{tabular}
    \caption{Comparing AutoCAD with IRM and C2L on NLI. We report the mean and the standard deviation over 5 random seeds.}
    \label{tab:morebaseline_nli}
\end{table}

\begin{table}[h]
    \centering
    \begin{tabular}{lcc}
    \toprule
    Method & Avg. \\
    \midrule
    Original & 86.43 \\
    StyleTransfer \cite{shen2017style} & 83.96 \\
    \midrule
    AutoCAD & \textbf{90.00}\\
    \bottomrule
    \end{tabular}
    \caption{Comparing AutoCAD with Sentiment Style Transfer on SST.}
    \label{tab:morebaseline_sst}
\end{table}

\begin{table*}[ht]
    \centering
    \small
    {
      \begin{tabular}{lccccccccc}
        \toprule
        \multirow{3}{*}{\textbf{Method}(Train loss)} & \multicolumn{1}{c}{\textbf{In-Domain}} & \multicolumn{2}{c}{\textbf{Out-of-Domain}} & \multicolumn{4}{c}{\textbf{Challenge}} & \multirow{3}{*}{\textbf{Avg. (\%   )}} \\
        \cmidrule(lr){2-2} \cmidrule(lr){3-4} \cmidrule(lr){5-8} 
        & SNLI & MNLI-m & MNLI-mm & CAD
        & Diagnostic & Stress & Break \\
        \midrule
        Original($\mathcal{L}_{ce}$)   & 84.84 & 63.02 & 63.84 & 61.25 & 50.27 & 54.55 & 69.32 & 60.38\\
        \midrule
        AutoCAD ($\mathcal{L}_{ce}$)    & \textbf{87.25} & \textbf{69.67} & \textbf{70.27} & \textbf{71.43} & \textbf{54.26} & \textbf{59.13} & \textbf{89.59} & \textbf{68.38} \\
        AutoCAD ($\mathcal{L}_{cf}$)    & 85.46 & 65.95 & 67.16 & 66.69 & 51.36 & 55.92 & 77.21 & 64.05\\
        \bottomrule
        \end{tabular}
    }
    \caption{Analysis of the effect of counterfactual loss for training classification models. Experiments are conducted on BERT\textsubscript{\tiny{BASE}}.}
    \label{tab:cf_loss}
\end{table*}

\subsection{Exploration of Robust Training for Classifier}

In our main experiment, we simply combine the data generated by AutoCAD with the original data and use cross-entropy loss to train the final classifier. Considering the generated data are not perfect and may conflict with the target labels, we wonder if adopting a finer training method can better utilize the heterogeneous data and further improve the task performance. Therefore, we additionally investigate the effectiveness of the counterfactual loss \cite{chang2021towards}, which does not require a label for the counterfactual data and is proved to work in training robust image classification models on CAD. Specifically, given $(\hat{x}, \hat{y})$ as a generated counterfactual from the original example $(x,y)$, we compare the following two losses:

\begin{equation}
    \begin{aligned}
    \mathcal{L}_{ce} &= -log(P(\hat{y}|\hat{x})) \\
    \mathcal{L}_{cf} &= -log(1-P(y|\hat{x}))
    \nonumber
\end{aligned}
\end{equation}
where $\mathcal{L}_{ce}$ is the standard cross-entropy loss and $\mathcal{L}_{cf}$ is the conterfactual loss. The cross-entropy loss on $(x,y)$ is omitted for simplicity.

We conduct experiments on SNLI with BERT\textsubscript{\tiny{BASE}}. The results are shown in Table~\ref{tab:cf_loss}. We find that while training with $\mathcal{L}_{cf}$ on data augmented by AutoCAD also brings substantial improvements, $\mathcal{L}_{ce}$ consistently and significantly outperforms $\mathcal{L}_{cf}$. We conjecture that the label noise introduced by AutoCAD is relatively low. Therefore, the cross-entropy loss, which provides a stronger supervised signal, can more fully exploit the augmented data to train a robust classification model. As we focus on automatically generating counterfactuals in this paper, we leave the exploration of how to train a better classification model with AutoCAD for future work.

\subsection{Generation Examples} \label{sec:examples}
We present detailed cases of the counterfactually augmented data generated by AutoCAD in Table~\ref{tab:snli_examples} and Table~\ref{tab:sst2_examples}. On the one hand, AutoCAD can identify valid rationales across different tasks. On the other hand, AutoCAD can generate authentic and diverse counterfactuals that conform to the target labels. Moreover, AutoCAD does not just apply simple rules to achieve label flipping. For example, in Table~\ref{tab:snli_examples}, AutoCAD generates ``\textit{young}'' against ``\textit{old}'' in (b) and ``\textit{crossing a river}'' against ``\textit{in the snow}'' in (c) as contradictions, instead of simply generating negation words such as ``\textit{not}'' or ``\textit{nobody}''. Another example shown in Table~\ref{tab:sst2_examples} (b) demonstrates that AutoCAD can generate counterfactuals with implicit negative expressions such as ``\textit{the only positive}''. Interestingly, we also find that AutoCAD can generate plausible counterfactuals even after misidentifying rationales (see Table~\ref{tab:sst2_examples} (c)).

\begin{table*}[t]
    \centering
    \subcaptionbox{}{
            \begin{tabular}{lp{14cm}}
            \toprule
            \midrule
            \multicolumn{2}{c}{\it{\textbf{Original}}}\\
            \midrule
            \textbf{Premise} & A man in an orange \hlgreen{kayak} and a man in a blue \hlgreen{kayak paddle} around a beautiful \hlgreen{lake}. \\
            \textbf{Hypothesis} & Two men are \hlgreen{sitting} in \hlgreen{front} of a \hlgreen{fire}. \\
            \textbf{Label} & \hlgreen{Contradiction} \\
            \midrule
            \midrule
            \multicolumn{2}{c}{\it{\textbf{AutoCAD}}}\\
            \midrule
            \textbf{Premise} & A man in an orange kayak and a man in a blue kayak paddle around a beautiful lake. \\
            \textbf{Hypothesis} & Two men are \hlred{kayaking} in the \hlred{middle} of a \hlred{lake}. \\
            \textbf{Label} & \hlred{Entailment} \\
            \midrule
            \textbf{Premise} & A man in an orange \hlred{jacket sits} and a man in a blue \hlred{jacket} around a beautiful \hlred{fire.} \\
            \textbf{Hypothesis} & Two men are sitting in front of a fire. \\
            \textbf{Label} & \hlred{Entailment} \\
            \midrule
            \textbf{Premise} & A man in an orange kayak and a man in a blue kayak paddle around a beautiful lake. \\
            \textbf{Hypothesis} & Two men are \hlyellow{kayak racing} in \hlyellow{front} of a \hlyellow{crowd}. \\
            \textbf{Label} & \hlyellow{Neutral} \\
            \bottomrule
            \end{tabular}
        }
    \subcaptionbox{}{
            \begin{tabular}{lp{14cm}}
            \toprule
            \midrule
            \multicolumn{2}{c}{\it{\textbf{Original}}}\\
            \midrule
            \textbf{Premise} & A \hlred{old} woman is \hlred{holding shopping bags} on a busy city street. \\
            \textbf{Hypothesis} & A \hlred{woman} on a busy street \hlred{holds some shopping bags}. \\
            \textbf{Label} & \hlred{Entailment} \\
            \midrule
            \midrule
            \multicolumn{2}{c}{\it{\textbf{AutoCAD}}}\\
            \midrule
            \textbf{Premise} & A \hlyellow{blond} woman is \hlyellow{chatting} on a busy city street. \\
            \textbf{Hypothesis} & A woman on a busy street holds some shopping bags \\
            \textbf{Label} & \hlyellow{Neutral} \\
            \midrule
            \textbf{Premise} & A old woman is holding shopping bags on a busy city street. \\
            \textbf{Hypothesis} & A \hlgreen{young} woman on a busy street \hlgreen{holding shopping bags}.\\
            \textbf{Label} & \hlgreen{Contradiction} \\
            \bottomrule
            \end{tabular}
        }
    \subcaptionbox{}{
            \begin{tabular}{lp{14cm}}
            \toprule
            \midrule
            \multicolumn{2}{c}{\it{\textbf{Original}}}\\
            \midrule
            \textbf{Premise} & Two \hlred{hikers crossing} a \hlred{snowy field}, with mountainous terrain behind them. \\
            \textbf{Hypothesis} & Two \hlred{hikers} are \hlred{out} in the \hlred{snow.} \\
            \textbf{Label} & \hlred{Entailment} \\
            \midrule
            \midrule
            \multicolumn{2}{c}{\it{\textbf{AutoCAD}}}\\
            \midrule
            \textbf{Premise} & Two \hlgreen{cyclists crossing} a \hlgreen{river}, with mountainous terrain behind them. \\
            \textbf{Hypothesis} & Two hikers are out in the snow. \\
            \textbf{Label} & \hlgreen{Contradiction} \\
            \midrule
            \textbf{Premise} & Two \hlgreen{cyclists  traverse} a \hlgreen{ridge}, with mountainous terrain behind them. \\
            \textbf{Hypothesis} & Two hikers are out in the snow. \\
            \textbf{Label} & \hlgreen{Contradiction} \\
            \bottomrule
            \end{tabular}
        }
\caption{Examples of original SNLI data and counterfactually augmented data generated by AutoCAD. We highlight the identified rationales in original examples and the generated spans by AutoCAD.}
\label{tab:snli_examples}
\end{table*}

\begin{table*}[t]
    \centering
    \subcaptionbox{}{
            \begin{tabular}{lp{15cm}}
            \toprule
            \midrule
            \multicolumn{2}{c}{\it{\textbf{Original}}}\\
            \midrule
            \textbf{Text} & Linklater fans , or pretentious types who want to appear avant-garde will \hlred{suck up} to this project ...\\
            \textbf{Label} & \hlred{Negative} \\
            \midrule
            \midrule
            \multicolumn{2}{c}{\it{\textbf{AutoCAD}}}\\
            \midrule
            \textbf{Text} & Linklater fans , or pretentious types who want to appear avant-garde will \hlgreen{adore} to this project ...\\
            \textbf{Label} & \hlgreen{Positive} \\
            \midrule
            \textbf{Text} & Linklater fans , or pretentious types who want to appear avant-garde will \hlgreen{definitely love} to this project ...\\
            \textbf{Label} & \hlgreen{Positive} \\
            \bottomrule
            \end{tabular}
        }
    \subcaptionbox{}{
            \begin{tabular}{lp{15cm}}
            \toprule
            \midrule
            \multicolumn{2}{c}{\it{\textbf{Original}}}\\
            \midrule
            \textbf{Text} & \hlgreen{The best thing} I can say about this film is that \hlgreen{I can't wait} to see what the \hlgreen{director} does next .\\
            \textbf{Label} & \hlgreen{Positive} \\
            \midrule
            \midrule
            \multicolumn{2}{c}{\it{\textbf{AutoCAD}}}\\
            \midrule
            \textbf{Text} & \hlred{The only positive thing} I can say about this film is that \hlred{I dread} to see what the \hlred{director} does next .\\
            \textbf{Label} & \hlred{Negative} \\
            \bottomrule
            \end{tabular}
        }
    \subcaptionbox{}{
            \begin{tabular}{lp{15cm}}
            \toprule
            \midrule
            \multicolumn{2}{c}{\it{\textbf{Original}}}\\
            \midrule
            \textbf{Text} & We have an \hlred{actor} who is great fun to watch \hlred{performing in a film that} is only mildly diverting.\\
            \textbf{Label} & \hlred{Negative} \\
            \midrule
            \midrule
            \multicolumn{2}{c}{\it{\textbf{AutoCAD}}}\\
            \midrule
            \textbf{Text} & We have an actor who is great fun to watch, \hlgreen{even if the movie} is only mildly diverting ..\\
            \textbf{Label} & \hlgreen{Positive} \\
            \midrule
            \textbf{Text} & We have an actor who is great fun to watch, \hlgreen{despite a script} is only mildly diverting ..\\
            \textbf{Label} & \hlgreen{Positive} \\
            \bottomrule
            \end{tabular}
        }
    \subcaptionbox{}{
            \begin{tabular}{lp{15cm}}
            \toprule
            \midrule
            \multicolumn{2}{c}{\it{\textbf{Original}}}\\
            \midrule
            \textbf{Text} & A \hlgreen{literary} detective story is still a detective story and aficionados of the \hlgreen{whodunit won't} be \hlgreen{disappointed}.\\
            \textbf{Label} & \hlgreen{Positive} \\
            \midrule
            \midrule
            \multicolumn{2}{c}{\it{\textbf{AutoCAD}}}\\
            \midrule
            \textbf{Text} & A \hlred{well-written, but uninspired} detective story is still a detective story and aficionados of the \hlred{genre will} be \hlred{disappointed}.\\
            \textbf{Label} & \hlred{Negative} \\
            \bottomrule
            \end{tabular}
        }
\caption{Examples of original SST-2 data and counterfactually augmented data generated by AutoCAD. We highlight the identified rationales in original examples and the generated spans by AutoCAD.}
\label{tab:sst2_examples}
\end{table*}
\balance

\end{document}